\documentclass[11pt,a4paper]{article}
\usepackage{authblk}  
\usepackage{mrp19}
\usepackage{times}
\usepackage{latexsym}
\usepackage{url}

\usepackage{tabularx}
\usepackage{caption}
\usepackage{multirow}
\usepackage{array}
\usepackage[pdftex]{graphicx}
\usepackage{amsmath}
\PassOptionsToPackage{normalem}{ulem}
\usepackage{ulem}
\usepackage{amssymb}
\usepackage{comment}
\usepackage{bm}
\usepackage{xspace}
\usepackage[inline]{enumitem}
\setlist{nosep}  
\setlist[description]{leftmargin=2ex}  
\setlist[enumerate]{labelindent=0ex,labelwidth=1.5ex,leftmargin=!}
\usepackage{colortbl}

\newcommand{\softmax}{\mathop{\rm softmax}\limits}
\global\long\def\T#1{#1^{\top}}

\DeclareMathAlphabet{\mathpzc}{OT1}{pzc}{m}{it}
\usepackage{mathtools}
\usepackage{here}

\usepackage{cleveref}
\crefname{section}{Section}{Sections}
\Crefname{section}{Section}{Sections}
\crefname{appendix}{Appendix}{Appendices}
\crefname{Appendix}{Appendix}{Appendices}
\crefname{subsection}{Section}{Sections}
\Crefname{subsection}{Section}{Sections}
\crefname{equation}{Equation}{Equations}
\Crefname{equation}{Equation}{Equations}
\Crefname{figure}{Figure}{Figures}
\crefname{figure}{Figure}{Figures}
\Crefname{table}{Table}{Tables}
\crefname{table}{Table}{Tables}
\crefname{enumi}{}{}
\usepackage{autonum}

\definecolor{azure}{rgb}{0.0, 0.5, 1.0}
\newcommand{\hyperparam}[1]{{\color{azure} \textsubscript{#1}}}
\definecolor{tbg}{gray}{0.9}
\newcommand{\sysrank}[1]{{\color{azure} /#1}}

\aclfinalcopy 

\title{\emph{Hitachi} at MRP~2019:\ {U}nified Encoder-to-Biaffine Network for Cross-Framework Meaning Representation Parsing}

\author[ ]{Yuta Koreeda\thanks{\quad Contributed equally.}}
\newcommand\CoAuthorMark{\footnotemark[\arabic{footnote}]} 
\author[ ]{Gaku Morio\protect\CoAuthorMark}
\author[ ]{Terufumi Morishita\protect\CoAuthorMark}
\author[ ]{Hiroaki Ozaki\protect\CoAuthorMark}
\author[ ]{Kohsuke Yanai}
\affil[ ]{Hitachi, Ltd.}
\affil[ ]{Research \& Development Group}
\affil[ ]{Kokubunji, Tokyo, Japan}
\affil[ ]{{\tt \{yuta.koreeda.pb, gaku.morio.vn, terufumi.morishita.wp, }\authorcr{\tt hiroaki.ozaki.yu, kohsuke.yanai.cs\}@hitachi.com}}
\date{}

\begin{document}

\maketitle
\begin{abstract}
This paper describes the proposed system of the \emph{Hitachi} team for the Cross-Framework Meaning Representation Parsing (MRP 2019) shared task.
In this shared task, the participating systems were asked to predict nodes, edges and their attributes for five frameworks, each with different order of ``abstraction'' from input tokens.
We proposed a unified encoder-to-biaffine network for all five frameworks, which effectively incorporates a shared encoder to extract rich input features, decoder networks to generate anchorless nodes in UCCA and AMR, and biaffine networks to predict edges.
Our system was ranked fifth with the macro-averaged MRP F1 score of 0.7604, and outperformed the baseline unified transition-based MRP.
Furthermore, post-evaluation experiments showed that we can boost the performance of the proposed system by incorporating multi-task learning, whereas the baseline could not.
These imply efficacy of incorporating the biaffine network to the shared architecture for MRP and that learning heterogeneous meaning representations at once can boost the system performance.

\end{abstract}

\section{Introduction}\label{sec:introduction}

This paper describes the proposed system of the \emph{Hitachi} team for the CoNLL 2019 Cross-Framework Meaning Representation Parsing (MRP 2019) shared task.
The goal of the task was to design a system that predicts sentence-level graph-based meaning representations in five frameworks, each with its specific linguistic assumptions.
The task was formulated as prediction of nodes, edges and their attributes from an input sentence (see \citet{Oep:Abe:Haj:19} for details).
The target frameworks were \begin{enumerate*}[label=(\arabic*)]
  \item DELPH-IN MRS Bi-Lexical Dependencies \citep[DM;][]{flickinger_2000,ivanova_2012},
  \item Prague Semantic Dependencies \citep[PSD;][]{hajic_2012,miyao_2014},
  \item Elementary Dependency Structures \citep[EDS;][]{oepen_2006},
  \item Universal Conceptual Cognitive Annotation framework \citep[UCCA;][]{abend_2013,hershcovich-etal-2017-transition}, and 
  \item Abstract Meaning Representation \citep[AMR;][]{banarescu_2013}
\end{enumerate*}.

In this work, we propose to unify graph predictions in all frameworks with a single encoder-to-biaffine network.
This objective was derived from our expectation that it would be advantageous if a single neural network can deal with all the frameworks, because it allows all frameworks to benefit from architectural enhancements and it opens up possibility to perform multi-task learning to boost overall system performance.
We argue that it is non-trivial to formulate different kinds of graph predictions as a single machine learning problem, since each framework has different order of ``abstraction'' from input tokens.
Moreover, such formulation has hardly been explored, with few exceptions including unified transition-based MRP \cite{hershcovich_2018}, to which we empirically show the superiority of our system (\cref{sec:experiments}).
We also present a multi-task variant of such system, which did not make it to the task deadline.

Our {\it non}-multi-task system obtained the fifth position in the formal evaluation.
We also evaluated the multi-task setup after the formal run, showing multi-task learning can yield an improvement in the performance.
This result implies learning heterogeneous meaning representations at once can boost the system performance.

\begin{figure*}
	\begin{center}
		\includegraphics[width=\linewidth]{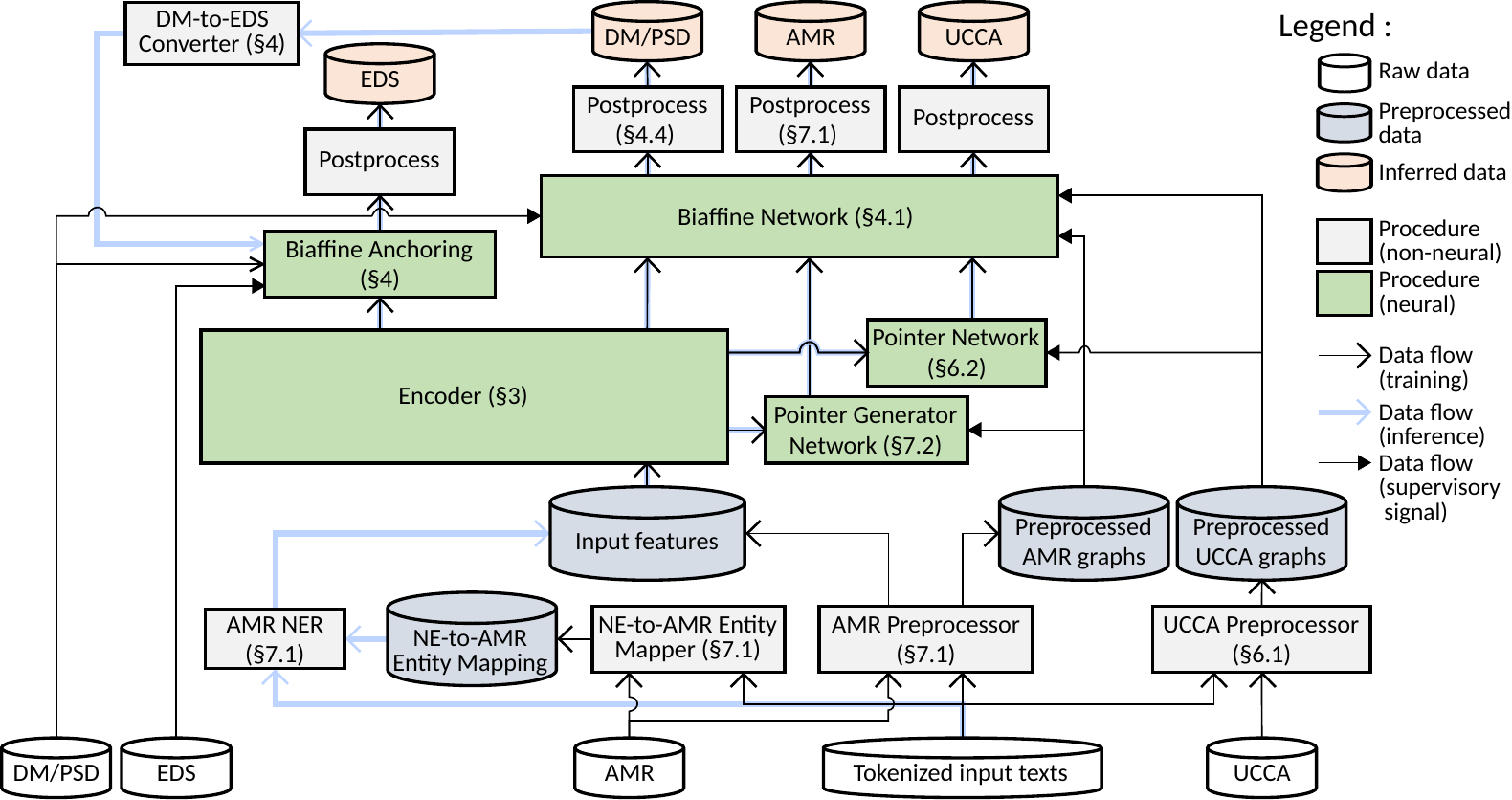}
		\caption{The overview of the proposed unified encoder-to-biaffine network for cross-framework meaning representation parsing.}
		\label{fig:overview}
	\end{center}
\end{figure*}

\section{Overview of the Proposed System}\label{sec:method}

The key challenge in unifying graph predictions with a single encoder-to-biaffine network lays in complementation of nodes, because the biaffine network can narrow down the node candidates but cannot generate new ones.
Our strategy is that we start from input tokens, generate missing nodes (nodes that do not have anchors to the input tokens) and finally predict edges with the biaffine network (\cref{fig:overview}).
More concretely, the shared encoder (\cref{sec:method-encoder}) fuses together rich input features for each token including features extracted from pretrained language models, which are then fed to bidirectional long short-term memories \citep[biLSTMs;][]{hochreiter_1997,schuster_1997} to obtain task-independent contextualized token representations.
The contextualized representations are fed to biaffine networks \cite{dozat_2018} to predict graphs for each framework along with the following framework-specific procedures:
\begin{description}
  \item[DM and PSD] Contextualized representations are fed to biaffine network to predict edges and their labels. They are also used to predict the node property \texttt{frame} (\cref{sec:method-sdp}).
  \item[EDS] The predicted DM graphs are converted to nodes and edges of EDS graphs. Contextualized representations are used to predict node anchors (\cref{sec:method-eds}). 
  \item[UCCA] Nodes in training data are serialized and aligned with input tokens. Contextualized representations are fed to a pointer network to generate non-terminal nodes, and to a biaffine network to predict edges and labels (\cref{sec:method-ucca}).
  \item[AMR] Contextualized representations are fed to pointer-generator network to generate nodes. Hidden states of the network are fed to a biaffine network to predict edges and their labels (\cref{sec:method-amr}). 
\end{description}

All models are trained end-to-end using mini-batch stochastic gradient decent with backpropagation (see \cref{sec:appendix-training-training} for the details).

\section{Shared Encoder}\label{sec:method-shared_encoder}

\subsection{Feature Extraction}\label{sec:method-feature_extractor}

Following work by \citet{dozat_2018} and \citet{zhang_2019}, we propose to incorporate multiple types of token representations to provide rich input features for each token.
Specifically, the proposed system combines surface, lemma, part-of-speech (POS) tags, named entity label, GloVe \cite{pennington_2014} embedding, ELMo \cite{peters_2018} embedding and BERT \cite{devlin_2019} embedding as input features .
The following descriptions explain how we acquire each input representations:
\begin{description}
  \item[Surface and lemma] We use the lower-cased node labels and the \texttt{lemma} properties from the companion data, respectively.
  Surfaces and lemmas that appear less than four times are replaced by a special \texttt{<UNK>} token.
  We also map numerical expressions\footnote{Surfaces or lemmas that can successfully be converted to numerics with \texttt{float} operation on Python 3.6} to a special \texttt{<NUM>} token.
  \item[POS tags] We use Universal POS tags and English specific POS tags from node properties \texttt{upos} and \texttt{xpos} in the companion data, respectively.
  \item[Named entity label] Named entity (NE) recognition is applied to the input text (see \cref{sec:method-amr-preprocessing}).
  \item[GloVe] We use 300-dimensional GloVe \cite{pennington_2014} pretrained on Common Crawl\footnote{\url{http://nlp.stanford.edu/data/glove.840B.300d.zip}} which are kept fixed during the training.
  Surfaces that do not appear in the pretrained GloVe are mapped to a special \texttt{<UNK>} token which is set to a vector whose values are randomly drawn from normal distribution with standard deviation of $1/\sqrt{\text{dimension of a GloVe vector}}$.
  \item[ELMo] We use the pretrained ``original'' ELMo\footnote{\url{https://s3-us-west-2.amazonaws.com/allennlp/models/elmo/2x4096_512_2048cnn_2xhighway/elmo_2x4096_512_2048cnn_2xhighway_weights.hdf5} and \url{elmo_2x4096_512_2048cnn_2xhighway_options.json}.}.
  Following \citet{peters_2018}, we ``mix'' different layers of ELMo for each token;
  \begin{align}
    \tilde{s_j}       &= \softmax_j(s_j) = \frac{\text{exp}(s_j)}{\sum_k{\text{exp}(s_k)}}, \\
    \mathbf{h}^{ELMo} &= \sum_{j = 0}^{N^{ELMo}-1}{\tilde{s_j}\mathbf{h}^{ELMo}_j},
  \end{align}
  where $\mathbf{h}^{ELMo}_j$ ($0 \leq j < N^{ELMo}$) is the hidden state of the $j$-th layer of ELMo, $\mathbf{h}^{ELMo}_0$ is the features from character-level CNN of ELMo, and $s_j$ are trainable parameters.
  $\mathbf{h}^{ELMo}_j$ are fixed in the training by truncating backpropagation to $\mathbf{h}^{ELMo}_j$.
  \item[BERT] We use the pretrained BERT-Large, Uncased (Original)\footnote{\url{https://s3.amazonaws.com/models.huggingface.co/bert/bert-large-uncased-pytorch_model.bin}, which is converted from the whitelisted BERT model in \url{https://github.com/google-research/bert}}.
  Since BERT takes subword units as input, a BERT embedding for a token is generated as the average of its subword BERT embeddings as in \citet{zhang_2019}.
\end{description}

The surface, lemma, POS tags and NE label of a token are each embedded as a vector.
The vectors are randomly initialized and updated during training.
To allow prediction of the top nodes for DM, PSD and UCCA, a special \texttt{<ROOT>} token is prepended to each input sequence.
For GloVe, ELMo and BERT, the \texttt{<ROOT>} is also embedded in the similar manner as other tokens with \texttt{<ROOT>} as the surface for the token.
A multi-layered perceptron (MLP) is applied to each of GloVe, ELMo and BERT embeddings.

To prevent the model from overrelying only on certain types of features, we randomly drop a group of features, where the groups are
\begin{enumerate*}[label=(\roman*)]
  \item lemma,
  \item POS tags and
  \item the rest.
\end{enumerate*}
All features in the same group are randomly dropped simultaneously but independently from other groups.

All seven features are then concatenated to form input token representation $\mathbf{h}^0_i$ (where $0 \leq i < L_{in}$ is the index of the token).

\subsection{Obtaining Contextualized Token Representation}\label{sec:method-encoder}

The input token representations $\mathbf{h}^0_i$ are fed to the multi-layered biLSTM with $N$ layers to obtain the contextualized token representations.
\begin{align}
    \overrightarrow{\mathbf{h}}^l_i &= \overrightarrow{\text{LSTM}}(\mathbf{h}^{l-1}_i, \overrightarrow{\mathbf{h}}^l_{i-1}, \overrightarrow{\mathbf{c}}^l_{i-1}),\\
	\overleftarrow{\mathbf{h}}^l_i &= \overleftarrow{\text{LSTM}}(\mathbf{h}^{l-1}_i, \overleftarrow{\mathbf{h}}^l_{i+1}, \overleftarrow{\mathbf{c}}^l_{i+1}),\\
    \mathbf{h}^l_i &= \left[\overrightarrow{\mathbf{h}}^l_i; \overleftarrow{\mathbf{h}}^l_i\right],
\end{align}
where $\mathbf{h}^l_i$ and $\mathbf{c}^l_i$ ($0 < l \leq N$) are the hidden states and the cell states of the $l$-th layer LSTM for $i$-th token.

\section{DM and PSD-specific Procedures}\label{sec:method-sdp}

\subsection{Biaffine Classifier}\label{sec:method-sdp-biaffine}

DM and PSD are Flavor (0) frameworks whose nodes have one-to-one correspondence to tokens.
We utilize biaffine networks to filter nodes, and to predict edges, edge labels and node attributes.
For each framework $\text{fw} \in \{\text{dm}, \text{psd}\}$, probability that there exists an edge $(i, j)$ from the $i$-th node to the $j$-th node $y_{\text{fw}, i, j}^\text{edge}$ is calculated for all pairs of nodes ($0 \leq i, j< L_{in}$).
\begin{equation}\label{eq:biaffine-edge}
\begin{aligned}
    \mathbf{h}_{\text{fw}, i}^\text{edge\_from} &= \text{MLP}^\text{edge\_from}(\mathbf{h}^N_i),\\
    \mathbf{h}_{\text{fw}, i}^\text{edge\_to} &= \text{MLP}^\text{edge\_to}(\mathbf{h}^N_i),\\
    y_{\text{fw}, i, j}^\text{edge} &= \sigma\left(\text{Biaff}^\text{edge}_\text{fw}\left(\mathbf{h}_{\text{fw}, i}^\text{edge\_from}, \mathbf{h}_{\text{fw}, j}^\text{edge\_to}\right)\right),
\end{aligned}
\end{equation}
where $\sigma$ is an element-wise sigmoid function.
Biaffine operation $\text{Biaff}^\text{edge}$ is defined as:
\begin{equation}
    \text{Biaff}^\text{edge}_\text{fw}\left(\mathbf{x}, \mathbf{y}\right) = \T{\mathbf{x}} \mathbf{U}^\text{edge}_\text{fw}\mathbf{y} + \mathbf{W}^\text{edge}_\text{fw}[\mathbf{x}; \mathbf{y}] + b^\text{edge}_\text{fw},
\end{equation}
where $\mathbf{U}^\text{edge}_\text{fw}$, $\mathbf{W}^\text{edge}_\text{fw}$ and $b^\text{edge}_\text{fw}$ are model parameters.
Probability of an edge $(i, j)$ being the $c$-th edge label $y_{\text{fw}, i, j, c}^\text{label}$ is calculated for all pairs of nodes.
\begin{equation}\label{eq:biaffine-edge_label}
\begin{aligned}
    \mathbf{h}_{\text{fw}, i}^\text{label\_from} &= \text{MLP}^\text{label\_from}(\mathbf{h}^N_i),\\
    \mathbf{h}_{\text{fw}, i}^\text{label\_to} &= \text{MLP}^\text{label\_to}(\mathbf{h}^N_i),\\
    t_{\text{fw}, i, j, c}^\text{label} &= \text{Biaff}_{\text{fw}, c}^\text{label}\left(\mathbf{h}_{\text{fw}, i}^\text{label\_from}, \mathbf{h}_{\text{fw}, j}^\text{label\_to}\right),\\
	y_{\text{fw}, i, j, c}^\text{label} &= \softmax_c\left(t_{\text{fw}, i, j, c}^\text{label}\right).
\end{aligned}
\end{equation}
Another form of biaffine operation for the edge label prediction $\text{Biaff}_{\text{fw}, c}^\text{label}$ is defined as:
\begin{equation}
    \text{Biaff}_{\text{fw}, c}^\text{label}\left(\mathbf{x}, \mathbf{y}\right) = \T{\mathbf{x}} \mathbf{U}^\text{label}_{\text{fw}, c}\mathbf{y} + \mathbf{W}^\text{label}_{\text{fw}, c}\mathbf{y},
\end{equation}
where $\mathbf{U}^\text{label}_{\text{fw}, c}$ and $\mathbf{W}^\text{label}_{\text{fw}, c}$ are model parameters.

A candidate edge $(i, j)$ whose edge probability $y_{\text{fw}, i, j}^\text{edge}$ ($0 < i, j$) exceeds $0.5$ is adopted as a valid edge.
Edge label with the highest probability $\arg\max_c y_{\text{fw}, i, j, c}$ is selected for each valid edge $(i, j)$.
A candidate top node $j$ whose edge probability $y_{\text{fw}, 0, j}^\text{edge}$ ($0 < j$) exceeds $0.5$ is adopted as a top node, allowing multiple tops.
Non-top nodes with no incoming or outgoing edge are discarded and remaining nodes are adopted as the predicted nodes.

\subsection{DM Frame Classifier}\label{sec:dm-class}

A DM node property \texttt{frame} consists of a frame type and frame arguments; e.g. \texttt{named:x-c} indicates the frame type is ``named entity'' with two possible arguments \texttt{x} and \texttt{c}.
The proposed system utilizes the contextualized features to predict the frame types and arguments separately.

Probability of the $i$-th node being $c$-th frame type $y_{\text{dm}, i, c}^\text{frame\_type}$ is predicted by applying MLP to the contextualized features:
\begin{align}
    t_{\text{dm}, i, c}^\text{frame\_type} &= \text{MLP}^\text{frame\_type}_c(\mathbf{h}^N_i),\\
    y_{\text{dm}, i, c}^\text{frame\_type} &= \softmax_c\left(t_{\text{dm}, i, c}^\text{frame\_type}\right).
\end{align}
The number of arguments for a frame is not fixed and the first argument can be trivially inferred from the frame type.
Thus, we predict from the second to the fifth arguments for each node.
Probability of $j$-th argument being $c$-th frame type $y_{\text{dm}, i, j, c}^\text{frame\_arg}$ is also predicted by applying MLP to the contextualized features:
\begin{align}
    t_{\text{dm}, i, j, c}^\text{frame\_arg} &= \text{MLP}^\text{frame\_arg}_{j, c}(\mathbf{h}^N_i),\\
    y_{\text{dm}, i, j, c}^\text{frame\_arg} &= \softmax_c\left(t_{\text{dm}, i, j, c}^\text{frame\_arg, j}\right).
\end{align}

\subsection{Training Objective}\label{sec:method-sdp-training}

DM and PSD are trained jointly in a multi-task learning setting but independently from other frameworks.
The loss for the edge prediction $\ell_\text{fw}^\text{edge}$ is cross entropy between the predicted edge $y_{i, j}^\text{edge}$ and the corresponding ground truth label.
A top node $j$ is treated as an edge $(0, j)$ and is trained along with the edge prediction.
The loss for the edge label prediction $\ell_\text{fw}^\text{label}$ is cross entropy between the predicted edge label $y_{i, j, c}^\text{label}$ and ground truth label.
The loss for the frame prediction $\ell_\text{dm}^\text{frame}$ is the sum of the frame type prediction loss $\ell_\text{dm}^\text{frame\_type}$ and the frame arguments prediction loss $\ell_\text{dm}^\text{frame\_arg}$, both of which are cross entropy loss between the prediction and the corresponding ground truth label.
Final multi-task loss is defined as:
\begin{equation}\label{eq:method-sdp-training-loss}
\begin{aligned}
	\ell_\text{sdp} =& \lambda^{\text{label}} \left(\ell^{\text{label}}_{\text{dm}} + \ell^{\text{label}}_{\text{psd}} + \lambda^{\text{frame}}\ell^{\text{frame}}_{\text{dm}}\right)\\
	      &+ \left(1 - \lambda^{\text{label}}\right)\left(\ell^{\text{edge}}_{\text{dm}} + \ell^{\text{edge}}_{\text{psd}}\right).
\end{aligned}
\end{equation}

\subsection{Postprocessing}\label{sec:method-sdp-postprocessing}

We reconstruct node property \texttt{frame} from the predicted frame types and arguments using external resources.
For DM, we filter out pairs of predicted frame type and arguments that do not appear in ERG SEM-I\footnote{\url{http://svn.delph-in.net/erg/tags/1214/etc}} or the training dataset (e.g. a word ``parse'' has only two possible frames \texttt{n:x} and \texttt{v:e-i-p}).
Then, we select a frame with the highest \emph{empirically scaled likelihood} which is calculated by scaling predicted joint probability $y_{\text{dm}, i, c}^\text{frame\_type}\prod_jy_{\text{dm}, i, j, c'}^\text{frame\_arg}$ proportionally to the frame frequency in the corpus.

For PSD, we use CzEngVallex\footnote{\url{http://hdl.handle.net/11234/1-1512}}, which contains frequency and the required arguments of each frame, to reconstruct frames.
We identify the frame type of a token from its lemma and POS tag.
Then, candidate frames are filtered using the required arguments (extracted by stripping \texttt{-suffix} from connected edges) and the most frequent frame is chosen as the node frame.

Token lemma is used for the node label, except for the special node labels in PSD  (e.g. \texttt{\#Bracket} and \texttt{\#PersPron}) that are looked-up from a hand-crafted dictionary using the surface and POS tag as a key.

\section{EDS-specific Procedure}\label{sec:method-eds}

DM graphs are constructed by lossy conversion from EDS graphs, both of which are derived from English Resource Semantics \citep[ERS;][]{flickinger-etal-2014-towards}.
Making use of such relationship, we developed heuristic inverse conversion from DM to EDS graphs by carefully studying EDS-to-DM conversion rules described in the ERG SEM-I corpus.
Specifically, our system generates EDS in three steps; the system \begin{enumerate*}[label=(\roman*)]
  \item convert all DM nodes to EDS surface nodes\footnote{For ease of explanation, we adopt a definition that ``the EDS surface nodes are the nodes that appear in DM and the abstract nodes are those that do not'' which results in slight inconsistence with the original definition.} with simple rules,
  \item\label{itm:eds-overview-gen} generate abstract nodes, and
  \item\label{itm:eds-overview-anchor} predict anchors for the abstract nodes.
\end{enumerate*}

\begin{figure}
	\begin{center}
		\includegraphics{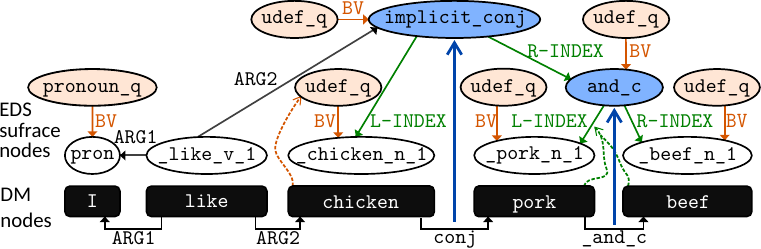}
		\caption{Generation of abstract nodes and their edges from \emph{I like chicken, pork and beef.}}
		\label{fig:eds-restoration}
	\end{center}
\end{figure}

We explain the generation of abstract nodes \cref{itm:eds-overview-gen} in details using an example in \cref{fig:eds-restoration}:
\begin{enumerate}
  \item Some abstract nodes (e.g. \texttt{\_and\_c}) and their node labels are generated with rules.
  \item\label{itm:eds-gen-detection} Presence of an abstract node on a node or an edge is detected with rules (e.g. \texttt{\_and\_c} implies presence of \texttt{\_q} node) or with binary logistic regression (e.g. \texttt{udef\_q} on \texttt{\_chicken\_n\_1}).
  \item The system predicts labels of the nodes generated in \cref{itm:eds-gen-detection} using multi-class logistic regression.
  \item The system predicts labels of edges from/to the generated nodes using multi-class logistic regression.
\end{enumerate}
POS tags, predicted DM frames and edge labels of adjacent nodes are used as features for the logistic regression.

We employ another neural network that utilize the contextualized features from the encoder to predict the anchors for the generated abstract nodes \cref{itm:eds-overview-anchor}.
For each abstract node (indexed $i$), let $\mathcal{T}_i$ be a subset of token indices $\mathcal{S} \equiv \{0,\dots,L_{in}-1\}$ each of which is selected as a DM node and the corresponding EDS surface node has the abstract node $i$ as an ancestor.
First, we create an input feature $x_{i,j}^\text{eds}$ ($j \in \mathcal{S}$) which is set as the label of node $i$ if $j \in \mathcal{T}_i$ or \texttt{<UNK>} otherwise.
Then, we embed $x^\text{eds}_{i,j}$ to obtain trainable vector $\mathbf{e}^\text{eds}_{i,j}$ and feed them to a biLSTM to obtain a contextualized representation $\bm{h}^\text{eds}_{i,j}$.
Finally, we predict a span in input tokens $[\text{argmax}_j y^\text{eds\_from}_{i,j}, \text{argmax}_j y^\text{eds\_to}_{i,j}]$ for the $i$-th abstract node,
\begin{equation}
\begin{aligned}
    y^\text{eds\_from}_{i,j} &= \softmax_j \left(\T{(\mathbf{h}^{\text{eds}}_{i,j})} \cdot \text{MLP}^\text{eds\_from}(\mathbf{h}^N_j)\right),\\
    y^\text{eds\_to}_{i,j} &= \softmax_j \left(\T{(\mathbf{h}^{\text{eds}}_{i,j})} \cdot \text{MLP}^\text{eds\_to}(\mathbf{h}^N_j)\right).
\end{aligned}
\end{equation}
The loss for the anchor prediction $\ell_\text{eds}$ is the sum of cross entropy between the predicted span $(y^\text{eds\_from}_{i,j}, y^\text{eds\_to}_{i,j})$ and the corresponding ground truth span.

\section{UCCA-specific Procedure}\label{sec:method-ucca}

\begin{figure}[tb]
  \centering
  \includegraphics{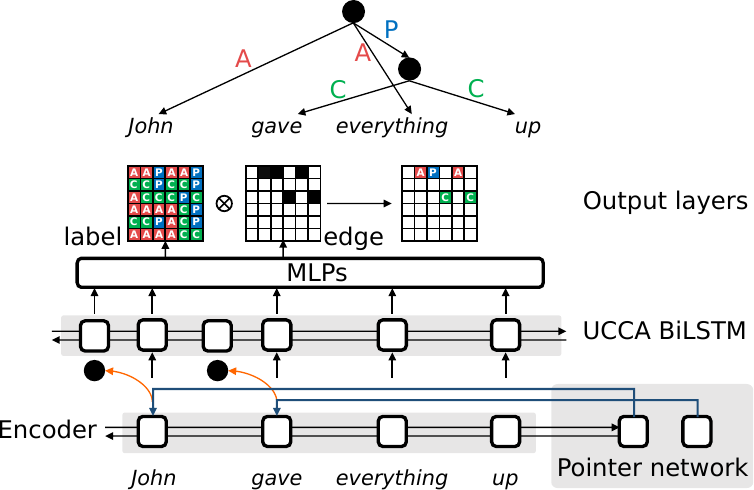}
  \caption{Illustration of UCCA parsing with pointer network and biaffine classifier.}
  \label{fig:ucca_model}
\end{figure}

A UCCA graph consists of terminal nodes which represent words, non-terminal nodes which represent internal structure, and labeled edges (e.g., participant (A), center (C), linker (L), process (P) and punctuation (U)) which represent connections between the nodes.
Motivated by the recent advances in constituency parsing, we predict spans of each terminal nodes at once without using any complicated mechanism as seen in transition-based \cite{Her:Arv:19} and greedy bottom-up \cite{yu-sagae-2019-uc} systems.
Our proposed UCCA parser (\cref{fig:ucca_model}) consists of \begin{enumerate*}[label=(\roman*)]
  \item a pointer network \cite{Vinyals_NIPS2015_5866} which generates non-terminal nodes from the contextualized token representations of the encoder,
  \item an additional biLSTM that encodes context of both the terminal and generated non-terminal nodes, and
  \item a biaffine network which predicts edges
\end{enumerate*}.

\subsection{Preprocessing} \label{sec:preprocessing-ucca}

We treat the generation of non-terminal nodes as a ``pointing'' problem.
Specifically, the system has to point the starting position of a span which has terminal or non-terminal children.
For example, upper part of \cref{fig:ucca_model} shows a graph with two non-terminal nodes $\bullet$.
The right non-terminal node has a span of {\it gave everything up}, and our system points at the starting position of the span {\it gave}.
By taking such strategy, we can serialize the graph in a consistent, straightforward manner; i.e. by inserting the non-terminal nodes to the left of the corresponding span.

The system also has to predict an anchor of a proper noun or a compound expression to merge constituent tokens into a single node.
For example, {\it no feathers in stock!!!!} is tokenized as ``{\it (no), (feathers), (in), (stock), (!), (!), (!), (!)}'' according to the companion data, but the UCCA parser is expected to output ``{\it (no), (feathers), (in), (stock), (!!!!)}''.
To solve the problem, we formulate the mergence of tokens as edge prediction; e.g. we assume that there exist virtual edges \texttt{CT} from leftmost constituent token to each subsequent token within a compound expression:
\vspace{-1em}
\begin{figure}[H]
\centering
\includegraphics[width=0.4\linewidth]{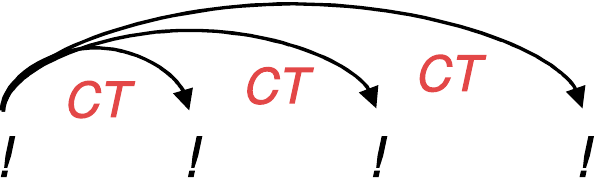}
\end{figure}
\vspace{-1em}

\noindent
and \texttt{CT} is predicted by the system along with the other edges.
There still exists tokenization discrepancy between the companion data and the graphs from EWT and Wiki.
The graphs with such discrepancy are simply discarded from the training data.

\subsection{Generating Non-terminal Nodes with Pointer Network}

Our system generates non-terminal nodes by pointing where to insert non-terminal nodes as described in \cref{sec:preprocessing-ucca}.
To point a terminal node, we employ a pointer network, which is a decoder that uses attention mechanism to produce probability distribution over the input tokens.
Given hidden states of the encoder $\mathbf{h}_{j}^N$, hidden states of the decoder are initialized by the last states of the shared encoder:
\begin{align}
    \mathbf{h}^\text{ucca\_dec}_{-1} &= \left[\overrightarrow{\mathbf{h}}^{N-K:N}_{L_{in}}; \overleftarrow{\mathbf{h}}^{N-K:N}_0\right],\\
  \mathbf{c}^\text{ucca\_dec}_{-1} &= \left[\overrightarrow{\mathbf{c}}^{N-K:N}_{L_{in}}; \overleftarrow{\mathbf{c}}^{N-K:N}_0\right],
\end{align}
where $K$ is the stacking number of the biLSTMs in the shared encoder.
We then obtain the hidden states of the decoder $\mathbf{h}^\text{ucca\_dec}_i$ as:
\begin{align}
  \mathbf{h}^\text{ucca\_dec}_i &= \text{LSTM}_\text{dec}(\mathbf{x}^\text{ucca\_dec}_i, \mathbf{h}^\text{ucca\_dec}_{i-1}, \mathbf{c}^\text{ucca\_dec}_{i-1}).
\end{align}
Attention distribution $\tilde{a}_{i,j}$ over the input tokens is calculated as:
\begin{align}
a_{i,j} &= \T{\mathbf{v}} \tanh \left(\mathbf{W}^\text{ucca\_dec} [\mathbf{h}^\text{ucca\_dec}_i;\mathbf{h}_{j}^N] \right),\\
\tilde{a}_{i,j} &= \softmax_j(a_{i,j}),
\end{align}
where $\mathbf{W}^\text{ucca\_dec}$ and $\mathbf{v}$ are parameters of the pointer network.
The successive input to the decoder $\mathbf{x}^\text{ucca\_dec}_{i+1}$ is the encoder states of the pointed token $\mathbf{h}^N_{\text{argmax}_j \tilde{a}_{i,j}}$.
$\mathbf{x}^\text{ucca\_dec}_i$ is chosen from the gold $\tilde{a}_{i,j}$ when training.

The decoder terminates its generation when it finally points the \texttt{<ROOT>}.
We obtain new hidden states $\mathbf{h}^\text{ucca\_ptr}_{i}$ ($0 \leq i \leq L_\text{ucca}$) by inserting pointer representations $\mathbf{h}^{\bullet}$ before the pointed token.
For example, {\it John gave everything up} (discussed above) will have hidden states
\begin{equation}
  \left(\mathbf{h}_\texttt{<ROOT>}^N, \mathbf{h}^{\bullet}, \mathbf{h}_\text{John}^N, \mathbf{h}^{\bullet}, \mathbf{h}_\text{gave}^N, \mathbf{h}_\text{everything}^N, \mathbf{h}_\text{up}^N \right).
\end{equation}
The pointer representation is defined as $\mathbf{h}^{\bullet} = \text{MLP}_{\bullet}(\mathbf{r})$, where $\mathbf{r}$ is a randomly initialized constant vector.

We note that the generated non-terminal nodes $\mathbf{h}^{\bullet}$ lack positional information because all $\mathbf{h}^{\bullet}$ have the same values.
To remedy this problem, a positional encoding \citet{Vaswani_NIPS2017_7181} is concatenated to each of $\mathbf{h}^\text{ucca\_ptr}_{i}$ to obtain position-aware $\mathbf{h}^\text{ucca\_ptr'}_{i}$.
Furthermore, we feed $\mathbf{h}^\text{ucca\_ptr'}_i$ to an additional biLSTM and obtain $\mathbf{h}^{\text{ucca}}_i$ in order to further encode the order information.

\subsection{Edge Prediction with Biaffine Network}

Now that we have contextualized representations for all candidate terminal and non-terminal nodes, the system can simply predict the edges and their labels in the exact same way as Flavor (0) graphs (\cref{sec:method-sdp-biaffine}).
Following \cref{eq:biaffine-edge} and \cref{eq:biaffine-edge_label}, we obtain probabilities if there exists an edge $(i, j)$, $y_{\text{ucca}, i, j}^\text{edge}$, and its label being $c$, $y_{\text{ucca}, i, j, c}^\text{label}$, with the input being $\mathbf{h}^{\text{ucca}}_i$ instead of $\mathbf{h}^N_i$.
We treat the remote edges\footnote{Edges for implicit relations and arguments. They were annotated as unlabeled edges each with an attribute \texttt{remote} in MRP.} independently but in the same way as the primary edges to predict $y_{\text{ucca}, i, j}^\text{remote}$.

The loss for the edge prediction $\ell_\text{ucca}^\text{edge}$, the edge label prediction $\ell^{\text{label}}_{\text{ucca}}$, the remote edge prediction $\ell^{\text{remote}}_{\text{ucca}}$ and the pointer prediction $\ell^{\text{dec}}_{\text{ucca}}$ are defined as cross entropy between the prediction $y_{\text{ucca}, i, j}^\text{edge}$, $y_{\text{ucca}, i, j, c}^\text{label}$, $y_{\text{ucca}, i, j}^\text{remote}$ and $\tilde{a}_{i,j}$ with the corresponding ground truth labels, respectively.
Thus, we arrive at the multi-task objective defined as:
\begin{equation}
  \begin{aligned}
    \ell_\text{ucca} =& \lambda^\text{edge}_\text{ucca} \ell^{\text{edge}}_{\text{ucca}} + \lambda^\text{label}_\text{ucca} \ell^{\text{label}}_{\text{ucca}} \\
    & + \lambda^\text{remote}_\text {ucca} \ell^{\text{remote}}_{\text{ucca}} + \lambda^\text{dec}_\text{ucca} \ell^{\text{dec}}_{\text{ucca}}.
  \end{aligned}
\end{equation}

\section{AMR-specific Procedures}\label{sec:method-amr}

Because AMR graphs do not have clear alignment between input tokens and nodes, the nodes have to be identified in prior to predicting edges.
Following \citet{zhang_2019}, we incorporate a pointer-generator network (i.e. a decoder with copy mechanisms) for the node generation and a biaffine network for the edge prediction.
There are two key preconditions in using a pointer-generator network; i.e. \begin{enumerate*}[label=(\roman*)]
  \item\label{itm:method-amr-precond1} node labels and input tokens share fair amount of vocabulary to allow copying a node from input tokens, and
  \item\label{itm:method-amr-precond2} graphs are serialized in a consistent, straightforward manner for it to be easily predicted by sequence generation
\end{enumerate*}.
To this end, we apply preprocessing to raw AMR graphs, train model to generate preprocessed graphs, and reconstruct final AMR graphs with postprocessing.

\subsection{Preprocessing}\label{sec:method-amr-preprocessing}

We modify the input tokens and the node labels to account for the precondition \cref{itm:method-amr-precond1}.
A node labeled with \texttt{.*-entity} or a subgraph connected with \texttt{name} edge is replaced with a node whose label is an anonymized entity label such as \texttt{PERSON.0} \cite{Konstas_2017}.
Then, for each entity node, a corresponding span of tokens is identified by rules similar to \citet{flanigan-etal-2014-discriminative}; i.e. a span of tokens with the longest common prefix between the token surfaces and the node attribute (e.g. for \texttt{date-entity} whose attribute \texttt{month} is \texttt{11}, we search for ``November'' and ``Nov'' in the token surfaces).
Unlike \citet{zhang_2019} which has replaced input token surfaces with anonymized entity labels, we add them as an additional input feature as described in \cref{sec:method-feature_extractor} to avoid hurting the performance of other frameworks.
At the prediction, we first identify NE tags in input tokens with Illinois NER tagger \cite{RatinovRo_2009}.
Then we map them to anonymized entity labels with frequency-based mapping constructed from the training dataset.

For non-entity nodes, we strip sense indices (e.g. \texttt{-01}) from node labels \cite{Lyu_2018}, which will then share fair amount of vocabulary with the input token lemmas.
Nodes with labels that still do not appear as lemmas after preprocessing are subject to normal generation from decoder vocabulary.

Directly serializing an AMR graph, which is a directed acyclic graph (DAG), may result in a complex conversion, which do not fulfill the precondition \cref{itm:method-amr-precond2}.
Therefore, we convert DAG to a spanning tree by replicating nodes with reentrancies (i.e. nodes with more than one incoming edge) for each incoming edge and serialize the graph with simple pre-order traversal over the tree.

\subsection{Extended Pointer-Generator Network}

We employ an extended pointer-generator network.
It automatically switches between three generation strategies; i.e. \begin{enumerate*}[label=(\arabic*)]
  \item \emph{source-side copy},
  \item \emph{decoder-side copy} that copies nodes that have been already generated, and
  \item normal generation from decoder vocabulary.
\end{enumerate*}
More formally, it uses attention mechanism to calculate probability distribution $\mathbf{p}_i$ over input tokens, generated nodes and node vocabulary.
Given contextualized token representation of the encoder $H^\text{enc}_l = \{\mathbf{h}_{0}^{l},\dots,\mathbf{h}_{L_{in}-1}^{l}\}$, we obtain hidden states of the decoder $\mathbf{h}^\text{amr}_i$ and $\mathbf{p}_i$ as:
\begin{equation}
\begin{aligned}
  \mathbf{h}^\text{amr}_i, \mathbf{p}_i &= \text{Decoder}_\text{amr}(\mathbf{h}^\text{enc'}_{i}, \mathbf{h}^\text{amr}_{i-1}, \mathbf{p}_{i-1}, H^\text{enc}_N),\\
  \mathbf{h}^\text{enc'}_i &= \text{Encoder}_\text{amr}(\mathbf{p}_i, \mathbf{h}^\text{amr}_0\dots\mathbf{h}^\text{amr}_{i-1}, H^\text{enc}_0),\\
  \mathbf{h}^\text{enc'}_0, \mathbf{h}^\text{amr}_{-1} &= \\
   \text{MLP}_\text{amr} & \left(\left[\overrightarrow{\mathbf{h}}^N_{L_{in}}; \overleftarrow{\mathbf{h}}^N_0;\overrightarrow{\mathbf{c}}^N_{L_{in}}; \overleftarrow{\mathbf{c}}^N_0\right]\right).
\end{aligned}
\end{equation}
$\text{Encoder}_\text{amr}$ treats a node as if it is a token, and utilizes the encoder (\cref{sec:method-shared_encoder}) with shared model parameters to obtain representation of $(i-1)$-th generated nodes ${h}^\text{enc'}_{i}$.
Concretely, $\text{Encoder}_\text{amr}$ combines lemma (corresponds to the node label), POS tags (only when copied from a token) and GloVe (from the node label) of a node, embeds each of them to a feature vector using the encoder and concatenates feature vectors to obtain ${h}^\text{enc'}_{i}$.

\subsection{Edge Prediction with Biaffine Network}

Now that we have representations $\mathbf{h}^\text{amr}_i$ for all nodes, the system can simply predict the edges and their labels in the same way as Flavor (0) graphs (\cref{sec:method-sdp-biaffine}).
Following \cref{eq:biaffine-edge} and \cref{eq:biaffine-edge_label}, we obtain probabilities that there exists an edge $(i, j)$, $y_{\text{amr}, i, j}^\text{edge}$, and its label being $c$, $y_{\text{amr}, i, j, c}^\text{label}$, with the input being $\mathbf{h}^{\text{amr}}_i$ instead of $\mathbf{h}^N_i$.
Note that we do not predict the top nodes for AMR, because the first generated node is always the top node in our formalism.

The loss for the edge prediction $\ell_\text{amr}^\text{edge}$, the edge label prediction $\ell^{\text{label}}_{\text{amr}}$,  and the decoder prediction $\ell^{\text{dec}}_{\text{amr}}$ are cross entropy between the prediction $y_{\text{amr}, i, j}^\text{edge}$, $y_{\text{amr}, i, j, c}^\text{label}$ and $\mathbf{p}_i$ with the corresponding ground truth labels, respectively.
Thus, we arrive at the multi-task loss for AMR defined as:
\begin{equation}
  \begin{aligned}
    \ell_\text{amr} =& \lambda^\text{biaf}_\text{amr}\left(\lambda^\text{label}_\text{amr} \ell^{\text{label}}_{\text{amr}} + (1 - \lambda^\text{label}_\text{amr}) \ell^{\text{edge}}_{\text{amr}}\right) \\
    & + \lambda^\text{cov}_\text{amr}\ell^\text{cov}_\text{amr} + (1 - \lambda^\text{biaf}_\text{amr} - \lambda^\text{cov}_\text{amr})\ell^{\text{dec}}_{\text{amr}},
  \end{aligned}
\end{equation}
where $\ell^\text{cov}_\text{amr}$ is coverage loss \cite{zhang_2019}.

For node prediction, we adopt beam search with search width of five.
For edge prediction, we apply Chu-Liu-Edmonds algorithm to find the maximum spanning tree.
Postprocessing, which includes inverse transformation of the preprocessing, is applied to reconstruct final AMR graphs.

\begin{table}[t]
    \centering
    \caption{MRP F1 scores for the formal run (shown as ``score\sysrank{rank}'')}\label{tab:experiments-leaderboard}
    \fontsize{8pt}{10pt}\selectfont
    \setlength{\tabcolsep}{1.3pt}
    \renewcommand{\arraystretch}{.8}
    \begin{tabular}{lllllll}\hline
        Team & Mean & DM & PSD & EDS & UCCA & AMR\\\hline
        HIT-SCIR & .8620\sysrank{1} & .9508\sysrank{2} & .9055\sysrank{4} & .9075\sysrank{2} & .8167\sysrank{1} & .7294\sysrank{2}\\
        SJTU-NICT & .8527\sysrank{2} & .9550\sysrank{1} & .9119\sysrank{3} & .8990\sysrank{3} & .7780\sysrank{3} & .7197\sysrank{3}\\
        SUDA-Alibaba & .8396\sysrank{3} & .9226\sysrank{7} & .8556\sysrank{8} & .9185\sysrank{1} & .7843\sysrank{2} & .7172\sysrank{5}\\
        Saarland & .8187\sysrank{4} & .9469\sysrank{4} & .9128\sysrank{1} & .8910\sysrank{4} & .6755\sysrank{6} & .6672\sysrank{6}\\
        \textbf{Hitachi} (ours) & \textbf{.7604\sysrank{5}} & \textbf{.9102\sysrank{8}} & \textbf{.9121\sysrank{2}} & \textbf{.8374\sysrank{6}} & \textbf{.7036\sysrank{5}} & \textbf{.4386\sysrank{8}}\\
        \'{U}FAL MRPipe & .7474\sysrank{6} & .8495\sysrank{9} & .7627\sysrank{9} & .6745\sysrank{7} & .7322\sysrank{4} & .7183\sysrank{4}\\
        ShanghaiTech & .6697\sysrank{7} & .9488\sysrank{3} & .8949\sysrank{6} & .8690\sysrank{5} & - & .6359\sysrank{7}\\
        Amazon & .5132\sysrank{8} & .9326\sysrank{6} & .8998\sysrank{5} & - & - & .7338\sysrank{1}\\
        JBNU & .4652\sysrank{9} & .9401\sysrank{5} & .8788\sysrank{7} & - & .5069\sysrank{7} & -\\
        SJTU & .4303\sysrank{10} & .4315\sysrank{11} & .4761\sysrank{11} & .5321\sysrank{8} & .3266\sysrank{9} & .3851\sysrank{9}\\
        \'{U}FAL-Oslo & .3442\sysrank{11} & .8051\sysrank{10} & .6092\sysrank{10} & .3064\sysrank{9} & - & -\\
        HKUST & .2450\sysrank{12} & .3699\sysrank{12} & .3529\sysrank{12} & - & .5021\sysrank{8} & -\\
        Bocharov & .0655\sysrank{13} & - & - & - & - & .3273\sysrank{10}\\\hline
        TUPA\textsuperscript{\textdagger} single & .5770 & .5554 & .5176 & .8100 & .2756 & .4473\\
        TUPA\textsuperscript{\textdagger} multi & .4534 & .4270 & .5265 & .7395 & .2365 & .3375\\\hline
        \multicolumn{7}{l}{\fontsize{6pt}{8pt}\selectfont\textsuperscript{\textdagger} baseline \cite{Her:Arv:19}}\\
    \end{tabular}
\end{table}

\section{Multi-task Variant}

We developed multi-task variant after the formal run.
Multi-task variant is trained to minimize following multi-task loss,
\begin{equation}\label{eq:method-mtl-loss}
\begin{aligned}
	\ell_{mt} =& \lambda^\text{biaf}\bigg(\begin{aligned}[t]
	           &\lambda^{\text{label}} \Big(\sum_\text{fw}\ell^{\text{label}}_{\text{fw}} + \lambda^\text{frame}\ell^\text{frame}_\text{dm}\Big)\\
	           & {} + \left(1 - \lambda^{\text{label}}\right)\sum_\text{fw}\ell^{\text{edge}}_{\text{fw}}\bigg) + \lambda^\text{cov}_\text{amr}\ell^\text{cov}_\text{amr}
	      \end{aligned}\\
          &{} + \sum_{\text{fw}\in\{\text{ucca},\text{amr}\}}\lambda^\text{dec}_\text{fw}\ell^{\text{dec}}_{\text{fw}} + \lambda^\text{remote}_\text{ucca} \ell^{\text{remote}}_{\text{ucca}}.\\
\end{aligned}
\end{equation}
All training data is simply merged and losses for frameworks that are missing in an input data are set to zero.
For example, if an input sentence has reference graphs for DM, PSD and AMR, losses for UCCA ($\ell^{\text{label}}_{\text{ucca}}$, $\ell^{\text{edge}}_{\text{ucca}}$, $\ell^{\text{dec}}_{\text{ucca}}$ and $\ell^{\text{remote}}_{\text{ucca}}$) are set to zero and sum of other losses are used to update the model parameters.
Then, the training data (sentences) are shuffled at the start of each epoch and are fed sequentially to update the model parameters as in normal mini-batch training.
No under-/over-sampling was done to scale the losses of frameworks, each with different number of reference graphs, but we instead applied early stopping for each framework separately (see \cref{sec:appendix-training} for the details).
For EDS, we do not train EDS anchor prediction jointly even in multi-task setting but apply transfer learning; the encoder of the EDS anchor prediction network is initialized from trained multi-task model.

We also experimented with a \emph{fine-tuned} multi-task variant.
For each target framework, we take the multi-task variant as a pretrained model (whose training data also includes the target framework) and train the model on the target framework independently to the other frameworks (except for DM and PSD, which are always trained together).

\section{Experiments}\label{sec:experiments}

\subsection{Method}

Experiments were carried out on the evaluation split of the dataset.
We applied hyperparameter tuning and ensembling to our system, which are detailed in \cref{sec:appendix-training} along with other training details.
BERT was excluded for the formal run since it did not make it to the task deadline.

We experimented with enhanced models with BERT after the formal run.
For these models, we adopted the best hyperparameters chosen by the submitted model without re-running the hyperparameter tuning.

All models were implemented using Chainer \cite{tokui_2015,akiba_2017}.

\subsection{Results}

\begin{table*}[t]
    \centering
    \caption{MRP and framework specific scores (shown as ``score\sysrank{rank}''). Gray background indicates that it is the score on LPPS subset.}\label{tab:experiments-details}
    \fontsize{8pt}{10pt}\selectfont
	\renewcommand{\arraystretch}{0.8}
    \begin{tabular}{lllllllll}\hline
		 & \multicolumn{7}{c}{MRP}  & Framework\\\cline{2-8}
		 Framework& Tops & Labels & Properties & Anchors & Edges & Attributes & All & specific\textsuperscript{\textdagger}\\\hline
		\multirow{2}{*}{All} & 0.8929\sysrank{3} & 0.6409\sysrank{6} & 0.5186\sysrank{9} & 0.7547\sysrank{5} & 0.6958\sysrank{5} & 0.0418\sysrank{7} & 0.7604\sysrank{5} & -\\
	     & \cellcolor{tbg}0.9167\sysrank{3} & \cellcolor{tbg}0.6238\sysrank{6} & \cellcolor{tbg}0.3743\sysrank{9} & \cellcolor{tbg}0.7602\sysrank{6} & \cellcolor{tbg}0.7025\sysrank{5} & \cellcolor{tbg}0.0340\sysrank{7} & \cellcolor{tbg}0.7618\sysrank{5} & \cellcolor{tbg}-\\
		\multirow{2}{*}{DM} & 0.9219\sysrank{6} & 0.9107\sysrank{6} & 0.8649\sysrank{9} & 0.9909\sysrank{4} & 0.9190\sysrank{5} & - & 0.9102\sysrank{9} & 0.9189\sysrank{5}\\
		& \cellcolor{tbg}0.9505\sysrank{5} & \cellcolor{tbg}0.8818\sysrank{8} & \cellcolor{tbg}0.8367\sysrank{10} & \cellcolor{tbg}0.9862\sysrank{6} & \cellcolor{tbg}0.9245\sysrank{5} & \cellcolor{tbg}- & \cellcolor{tbg}0.8939\sysrank{9} & \cellcolor{tbg}0.9272\sysrank{4}\\
		\multirow{2}{*}{PSD} & 0.9538\sysrank{5} & 0.9494\sysrank{3} & 0.9118\sysrank{7} & 0.9896\sysrank{5} & 0.7948\sysrank{5} & - & 0.9121\sysrank{2} & 0.8085\sysrank{4}\\
		 & \cellcolor{tbg}0.9515\sysrank{5} & \cellcolor{tbg}0.9204\sysrank{2} & \cellcolor{tbg}0.8366\sysrank{8} & \cellcolor{tbg}0.9820\sysrank{6} & \cellcolor{tbg}0.7846\sysrank{4} & \cellcolor{tbg}- & \cellcolor{tbg}0.8840\sysrank{2} & \cellcolor{tbg}0.8075\sysrank{4}\\
		\multirow{2}{*}{EDS} & 0.7319\sysrank{9} & 0.8225\sysrank{7} & 0.5851\sysrank{7} & 0.8694\sysrank{6} & 0.8497\sysrank{7} & - & 0.8374\sysrank{7} & 0.7826\sysrank{7}\\
		 & \cellcolor{tbg}0.8515\sysrank{7} & \cellcolor{tbg}0.7763\sysrank{7} & \cellcolor{tbg}0.0670\sysrank{9} & \cellcolor{tbg}0.8737\sysrank{7} & \cellcolor{tbg}0.8427\sysrank{7} & \cellcolor{tbg}- & \cellcolor{tbg}0.8110\sysrank{7} & \cellcolor{tbg}0.7571\sysrank{7}\\
		\multirow{2}{*}{UCCA} & 0.9965\sysrank{2} & - & - & 0.9238\sysrank{6} & 0.5588\sysrank{6} & 0.2092\sysrank{7} & 0.7036\sysrank{6} & 0.4277\sysrank{6}\\
		 & \cellcolor{tbg}0.9900\sysrank{2} & \cellcolor{tbg}- & \cellcolor{tbg}- & \cellcolor{tbg}0.9593\sysrank{7} & \cellcolor{tbg}0.6050\sysrank{6} & \cellcolor{tbg}0.1698\sysrank{7} & \cellcolor{tbg}0.7498\sysrank{6} & \cellcolor{tbg}0.5024\sysrank{6}\\
		\multirow{2}{*}{AMR} & 0.8604\sysrank{3} & 0.5221\sysrank{8} & 0.2314\sysrank{9} & - & 0.3568\sysrank{8} & - & 0.4386\sysrank{8} & 0.4254\sysrank{8}\\
		 & \cellcolor{tbg}0.8400\sysrank{4} & \cellcolor{tbg}0.5404\sysrank{8} & \cellcolor{tbg}0.1311\sysrank{9} & \cellcolor{tbg}- & \cellcolor{tbg}0.3558\sysrank{8} & \cellcolor{tbg}- & \cellcolor{tbg}0.4701\sysrank{8} & \cellcolor{tbg}0.4530\sysrank{8}\\\hline
		\multicolumn{9}{l}{\fontsize{6pt}{8pt}\selectfont\textsuperscript{\textdagger} DM/PSD: SDP labeled F1, EDS: EDM all F1, UCCA:UCCA labeled all F1, AMR: SMATCH F1}\\
    \end{tabular}
\end{table*}

\begin{table*}[t]
    \centering
    \caption{MRP F1 scores for the variants of the proposed system (shown as ``score\sysrank{rank}'' where the rank is calculated by assuming that it was the submitted model).}\label{tab:experiments-variants}
    \fontsize{8pt}{10pt}\selectfont
	\renewcommand{\arraystretch}{0.8}
    \begin{tabular}{lllllll}\hline
		Variant & Average & DM & PSD & EDS & UCCA & AMR\\\hline
		\texttt{SFL} & 0.7575\sysrank{5} & 0.9071\sysrank{9} & 0.9064\sysrank{3} & 0.8339\sysrank{7} & 0.7014\sysrank{6} & 0.4386\sysrank{8}\\
		\texttt{SFL(ensemble)}\textsuperscript{\textdagger} & 0.7604\sysrank{5} & 0.9102\sysrank{9} & 0.9121\sysrank{2} & 0.8374\sysrank{7} & 0.7036\sysrank{6} & 0.4386\sysrank{8}\\\hline
		\texttt{BERT+SFL(NT)} & 0.7450\sysrank{6} & 0.9038\sysrank{9} & 0.9069\sysrank{3} & 0.8301\sysrank{7} & 0.6945\sysrank{6} & 0.3896\sysrank{8}\\
		\texttt{BERT+MTL(NT)} & 0.7144\sysrank{6} & 0.8726\sysrank{9} & 0.8791\sysrank{7} & 0.7987\sysrank{7} & 0.6422\sysrank{6} & 0.3794\sysrank{9}\\
		\texttt{BERT+MTL+FT(NT)} & 0.7507\sysrank{5} & 0.9045\sysrank{9} & 0.9054\sysrank{4} & 0.8304\sysrank{7} & 0.7126\sysrank{6} & 0.4008\sysrank{8}\\\hline
		\multicolumn{7}{l}{\fontsize{6pt}{8pt}\selectfont \texttt{SFL}: single-framework learning, \texttt{MTL}: multi-task learning, \texttt{FT}: fine-tuning, \texttt{ensemble}: with ensembles,}\\
		\multicolumn{7}{l}{\fontsize{6pt}{8pt}\selectfont \texttt{NT}: random seed is not tuned, \textsuperscript{\textdagger} formal run}\\
    \end{tabular}
\end{table*}

The official results are shown in \cref{tab:experiments-leaderboard} and \cref{tab:experiments-details}.
Our system obtained macro-averaged MRP F1 score of 0.7604 and was ranked fifth amongst all submissions.
Our system outperformed conventional unified architecture for MRP \citep[TUPA baselines;][]{Her:Arv:19} in all frameworks but AMR.
This indicates the efficacy of using the biaffine network as a shared architecture for MRP.

Our system obtained relatively better (second) position in PSD.
This was due to relatively good performance on the node label prediction where we carefully constructed postprocessing rule for special nodes' labels (\cref{sec:method-sdp-postprocessing}) instead of just using lemmas.

Our system obtained significantly worse result in AMR (difference of 0.2952 MRP F1 score to the best performing system), even though our system incorporates the state-of-the-art AMR parser \cite{zhang_2019}.
One reason is that \citet{zhang_2019} was obtaining a large score boost from the Wikification task, which was not part of the MRP 2019 shared task.
Another reason could be that we may have missed out important implementation details for the pointer-generator network, since the implementation of \citet{zhang_2019} was not yet released at the time of our system development.

\Cref{tab:experiments-variants} shows the performance of other variants of the proposed system.
The single-framework learning variant (SFL) without BERT (\texttt{SFL}) performed better than SFL with BERT (\texttt{BERT+SFL(NT)}), which suggests that impact of hyperparameter tuning was larger than that of incorporating BERT.
The multi-task learning variant (MTL) with fine-tuning (\texttt{BERT+MTL+FT(NT)}) outperformed the SFL in the comparable condition (\texttt{BERT+SFL(NT)}).
This result implies learning heterogeneous meaning representations at once can boost the system performance.

\section{Conclusions}\label{sec:conclusion}

In this paper, we described our proposed system for the CoNLL 2019 Cross-Framework Meaning Representation Parsing (MRP 2019) shared task.
Our system was the unified encoder-to-biaffine network for all five frameworks.
The system was ranked fifth in the formal run of the task, and outperformed the baseline unified transition-based MRP.
Furthermore, post-evaluation experiments showed that we can boost the performance of the proposed system by incorporating multi-task learning.
These imply efficacy of incorporating the biaffine network to the shared architecture for MRP and that learning heterogeneous meaning representations at once can boost the system performance.

While our architecture successfully unified graph predictions in the five frameworks, it is non-trivial to extend the architecture to another framework.
It is because there could be a more suitable node generation scheme for a different framework and naively applying the pointer network for partial nodes complementation (or extended pointer-generator network for full nodes generation) may result in a poor performance.
Thus, it is our future work to design a more universal method for the node generation.

\bibliography{main}
\bibliographystyle{acl_natbib}

\appendix

\section{Training Details}\label{sec:appendix-training}

We split dataset into training dataset which was used to update model parameters, validation dataset (i) which was used for early stopping, and validation dataset (ii) which was used for hyperparameter tuning and construction of ensembles.
For AMR and UCCA, we selected sentences that appear in more than one framework to populate the training dataset, and extracted 500 (300) and 1500 (700) data from the rest as validation dataset (i) and (ii) for AMR (UCCA), respectively.
For DM, PSD and EDS, we selected data that appear in AMR or UCCA to populate the training dataset, and extracted 500 and 1500 data from the rest as validation dataset (i) and (ii), respectively.

\subsection{Model Training}\label{sec:appendix-training-training}

All models are trained using mini-batch stochastic gradient decent with backpropagation.
We use Adam optimizer \cite{kingma_2014} with gradient clipping.

For the non-multi-task variant, early stopping is applied for each framework with SDP labeled dependency F1 score \cite{oepen_2014} (for DM, PSD and UCCA) or validation loss (for EDS and AMR) as the objective.
Note that early stopping is applied separately to each framework for the joint training of DM and PSD.
Concretely, for the joint training of DM and PSD, we train the model with respect to the joint loss $\ell_\text{sdp}$ in \cref{eq:method-sdp-training-loss} but we use a model at a training epoch whose DM-specific (or PSD-specific) SDP labeled dependency F1 score is highest for DM (or PSD) prediction.

For the multi-task variants, we employ a slightly different strategy for early stopping.
For the multi-task variant without fine-tuning, we apply early stopping separately to each framework with respect to the framework-specific validation loss.
For example, we train the multi-task model with respect to $\ell_\text{mtl}$ in \cref{eq:method-mtl-loss} but we use a model at a training epoch whose PSD-specific validation loss $\lambda^{\text{label}}\ell^{\text{label}}_{\text{psd}} + \left(1 - \lambda^{\text{label}}\right)\ell^{\text{edge}}_{\text{psd}}$ is lowest for PSD prediction.
For each framework in the fine-tuned multi-task variant, we adopt the multi-task pretrained model at a training epoch whose framework-specific validation loss is lowest and fine-tune on the model in the same manner as the non-multi-task variant.
Note that, for DM and PSD, which are fine-tuned together even in the fine-tuned multi-task variant, we adopt the multi-task pretrained model at a training epoch whose multi-task validation loss $\ell_\text{mtl}$ is lowest.

Dropout \cite{srivastava_15} is applied to \begin{enumerate*}[label=(\alph*)]
  \item the input to each layer of the shared encoder,
  \item the input to the biaffine networks, and
  \item the input to each layer of the UCCA and AMR decoders
\end{enumerate*}.

\subsection{Hyperparameter Tuning}

We random searched subset of hyperparameters for DM, PSD, UCCA and AMR.
See \cref{tab:method-hyperparameters} for hyperparameter search space and the list of hyperparameters chosen by the best performing model in each framework.
We tried 20 hyperparameter sets for DM/PSD, 50 for UCCA, and 25 for AMR.

We did not tune the hyperparameters of the multi-task variants.
We adopted the best hyperparameters chosen in the non-multi-task variants (\cref{tab:method-hyperparameters}) and hand-tuned the hyperparameters by examining learning curves over few runs.
For the fine-tuning, we adopted the best hyperparameters chosen in the non-multi-task variants (\cref{tab:method-hyperparameters}).
See \cref{tab:method-hyperparameters-mtl} for the list of hyperparameters used in the multi-task variants.

\begin{table}[t!]
    \centering
    \caption{List of hyperparameters. Multiple values indicates that the hyperparameter was tuned within that values. Subscript d (DM), p (PSD), u (UCCA) and a (AMR) denotes the hyperparameter chosen by the best performing model on validation dataset. $\mathcal{U}(a, b)$ is a uniform distribution in $[a, b]$.}\label{tab:method-hyperparameters}
    \fontsize{8pt}{10pt}\selectfont
    \setlength{\tabcolsep}{2pt}
    \renewcommand{\arraystretch}{0.8}
    \begin{tabular}{lll}\hline
    \multicolumn{2}{l}{Hyperparameter} & Value or search space\\\hline
    \multicolumn{2}{l}{\textbf{Common}} & \\
	& Word embedding dimension & 100\\
	& Lemma embedding dimension & 100\\
	& POS embedding dimension & 100\\
	& NE embedding dimension & 100\\
	& GloVe MLP hidden size & 125\\
	& ELMo MLP hidden size & 512\\
    & Word drop probability & 0.1\hyperparam{dpua}, 0.2, 0.4\\
    & POS drop probability & 0.1\hyperparam{du}, 0.2\hyperparam{a}, 0.4\hyperparam{p}\\
    & Lemma drop probability & 0.1\hyperparam{p}, 0.2\hyperparam{da}, 0.4\hyperparam{u}\\
    & \# of layers in encoder & 2\hyperparam{pu}, 3\hyperparam{da}\\
    & Encoder LSTM hidden size & 256, 512\hyperparam{dpua}\\
    & Encoder dropout rate & 0.1\hyperparam{a}, 0.25\hyperparam{d}, 0.5\hyperparam{pu}\\
    & Biaffine input dropout & 0.2\hyperparam{pua}, 0.45\hyperparam{d}\\
    & Edge prediction dropout & 0.25\hyperparam{dpua}, 0.4\\
    & Learning rate & $10^{\mathcal{U}(-3.32, -2.92)}$\\
	&               & $\rightarrow$ 0.000858\hyperparam{d}, 0.000675\hyperparam{p},\\
	&               & \quad         0.00117\hyperparam{u}, 0.00059\hyperparam{a}\\
    & Adam $(\beta_1, \beta_2)$\textsuperscript{\textdagger} & (0.9, 0.999)\hyperparam{dp}, (0, 0.95)\hyperparam{ua}\\
    \multicolumn{2}{l}{\textbf{DM/PSD}} & \\
    & Edge MLP hidden size & 600\\
    & Edge label MLP hidden size & 600\\
    & Frame prediction MLP hidden size & 600\\
    & Frame prediction dropout & 0.2, 0.55\hyperparam{dp}\\
    & Edge label prediction dropout & 0.33\hyperparam{d}, 0.5\hyperparam{p}\\
    & Loss coefficient $\lambda^{\text{label}}_\text{fw}$ & $\mathcal{U}(0.02, 0.03)$\\
	&     & $\rightarrow$ 0.0210\hyperparam{d}, 0.0242\hyperparam{p}\\
    & Loss coefficient $\lambda^{\text{frame}}_\text{fw}$ & 0.5\\
    & \# of epochs & 50\\
    & Batch size & 64\\
    \multicolumn{2}{l}{\textbf{UCCA}} & \\
    & Edge MLP hidden size & 400, 500\hyperparam{u}, 600\\
    & Edge label MLP hidden size & 400\hyperparam{u}, 500, 600\\
    & Edge label prediction dropout & 0.25\hyperparam{u}, 0.33\\
    & Decoder dropout & 0.5\\
    & Loss coefficient $\lambda^\text{edge}_\text{ucca}$ & 0.3\\
    & Loss coefficient $\lambda^\text{label}_\text{ucca}$ & 0.3\\
    & Loss coefficient $\lambda^\text{remote}_\text{ucca}$ & 0.2\\
    & Loss coefficient $\lambda^\text{dec}_\text{ucca}$ &　0.2\\
    & \# of epochs & 40\\
    & Batch size & 100\\
    \multicolumn{2}{l}{\textbf{AMR}} & \\
    & Edge MLP hidden size & 600\\
    & Edge label MLP hidden size & 600\\
    & Edge label prediction dropout & 0.33\hyperparam{a}, 0.5\\
    & Decoder type\textsuperscript{\textdaggerdbl} & deep small\hyperparam{a}, shallow wide\\
    & Decoder dropout & 0.25, 0.33\hyperparam{a}, 0.5\\
    & Loss coefficient $\lambda^{\text{label}}_\text{amr}$ & $\mathcal{U}(0.1, 0.5)$ $\rightarrow$ 0.395\hyperparam{a}\\
    & Loss coefficient $\lambda^{\text{cov}}_\text{amr}$ & $\mathcal{U}(0.2, 0.4)$ $\rightarrow$ 0.339\hyperparam{a}\\
    & Loss coefficient $\lambda^{\text{gen}}_\text{amr}$ & $\mathcal{U}(0.2, 0.4)$ $\rightarrow$ 0.271\hyperparam{a}\\
    & \# of epochs & 50\\
    & Batch size & 64\\
    \hline
    \multicolumn{3}{l}{\fontsize{6pt}{8pt}\selectfont\textsuperscript{\textdagger} Commonly used setting and the setting used in \citet{dozat_2018}.}\\
    \multicolumn{3}{p{\dimexpr\linewidth-2\tabcolsep\relax}}{\fontsize{6pt}{8pt}\selectfont\textsuperscript{\textdaggerdbl}``deep small'' is three-layered LSTM with hidden size of 512 and ``shallow wide'' is two-layered LSTM with hidden size of 1024.}\\
    \end{tabular}
\end{table}

\begin{table}[t!]
    \centering
    \caption{Hyperparameters for the multi-task variants}\label{tab:method-hyperparameters-mtl}
    \fontsize{8pt}{10pt}\selectfont
    \setlength{\tabcolsep}{2pt}
    \renewcommand{\arraystretch}{0.8}
    \begin{tabular}{llll}\hline
    \multicolumn{3}{l}{Hyperparameter} & Value\\\hline
    \multicolumn{4}{l}{\textbf{Model architecture}} \\
    & & Word embedding dimension & 100\\
    & & Lemma embedding dimension & 100\\
    & & POS embedding dimension & 100\\
    & & NE embedding dimension & 100\\
    & & GloVe MLP hidden size & 125\\
    & & ELMo MLP hidden size & 512\\
    & & \# of layers in encoder & 3\\
    & & Encoder LSTM hidden size & 512\\
    & & Edge MLP hidden size & 600\\
    & & Edge label MLP hidden size & 600\\
    & & Frame prediction MLP hidden size & 600\\
    & & AMR decoder type\textsuperscript{\textdagger} & deep small\\
    \multicolumn{4}{l}{\textbf{Training conditions}} \\
    & \multicolumn{2}{l}{\textbf{Multi-task (pre)training}} & \\
    & & Word drop probability & 0.2\\
    & & POS drop probability & 0.2\\
    & & Lemma drop probability & 0.2\\
    & & Encoder dropout rate & 0.5\\
    & & Biaffine input dropout & 0.45\\
    & & Edge prediction dropout & 0.25\\
    & & Edge label prediction dropout & 0.33\\
    & & Learning rate & 0.00006\\
    & & Adam $(\beta_1, \beta_2)$\textsuperscript{\textdagger} & (0.9, 0.999)\\
    & & Loss coefficient $\lambda^\text{biaf}$ & 1.0\\
    & & Loss coefficient $\lambda^\text{label}$ & 0.15\\
    & & Loss coefficient $\lambda^\text{frame}$ & 0.5\\
    & & Loss coefficient $\lambda^\text{remote}_\text{ucca}$ & 0.5\\
    & & Loss coefficient $\lambda^\text{dec}_\text{ucca}$ &　0.08\\
    & & Loss coefficient $\lambda^\text{dec}_\text{amr}$ &　1.2\\
    & & Loss coefficient $\lambda^\text{cov}_\text{amr}$ & 1.0\\
    & & \# of epochs & 60\\
    & & Batch size & 128\\
    & \multicolumn{2}{l}{\textbf{DM/PSD fine-tuning}} & \\
    & & Word drop probability & 0.1\\
    & & POS drop probability & 0.2\\
    & & Lemma drop probability & 0.2\\
    & & Encoder dropout rate & 0.25\\
    & & Biaffine input dropout & 0.45\\
    & & Edge prediction dropout & 0.25\\
    & & Learning rate & 0.001\textsuperscript{\textdaggerdbl}\\
    & & Adam $(\beta_1, \beta_2)$\textsuperscript{\textdagger} & (0, 0.95)\textsuperscript{\textdaggerdbl}\\
    & & Frame prediction dropout & 0.55\\
    & & Edge label prediction dropout & 0.33\\
    & & Loss coefficient $\lambda^{\text{label}}_\text{fw}$ & 0.025\\
    & & Loss coefficient $\lambda^{\text{frame}}_\text{fw}$ & 0.5\\
    & & \# of epochs & 50\\
    & & Batch size & 64\\
    & \multicolumn{2}{l}{\textbf{UCCA fine-tuning}} & \\
    & & Word drop probability & 0.1\\
    & & POS drop probability & 0.1\\
    & & Lemma drop probability & 0.4\\
    & & Encoder dropout rate & 0.5\\
    & & Biaffine input dropout & 0.2\\
    & & Edge prediction dropout & 0.25\\
    & & Learning rate & 0.00117\\
    & & Adam $(\beta_1, \beta_2)$\textsuperscript{\textdagger} & (0, 0.95)\\
    & & Edge label prediction dropout & 0.25\\
    & & Decoder dropout & 0.5\\
    & & Loss coefficient $\lambda^\text{edge}_\text{ucca}$ & 0.3\\
    & & Loss coefficient $\lambda^\text{label}_\text{ucca}$ & 0.3\\
    & & Loss coefficient $\lambda^\text{remote}_\text{ucca}$ & 0.2\\
    & & Loss coefficient $\lambda^\text{dec}_\text{ucca}$ &　0.2\\
    & & \# of epochs & 40\\
    & & Batch size & 100\\
    & \multicolumn{2}{l}{\textbf{AMR fine-tuning}} & \\
    & & Word drop probability & 0.1\\
    & & POS drop probability & 0.2\\
    & & Lemma drop probability & 0.2\\
    & & Encoder dropout rate & 0.1\\
    & & Biaffine input dropout & 0.2\\
    & & Edge prediction dropout & 0.25\\
    & & Learning rate & 0.00059\\
    & & Adam $(\beta_1, \beta_2)$\textsuperscript{\textdagger} & (0, 0.95)\\
    & & Edge label prediction dropout & 0.33\\
    & & Decoder dropout & 0.33\\
    & & Loss coefficient $\lambda^{\text{label}}_\text{amr}$ & 0.395\\
    & & Loss coefficient $\lambda^{\text{cov}}_\text{amr}$ & 0.339\\
    & & Loss coefficient $\lambda^{\text{gen}}_\text{amr}$ & 0.271\\
    & & \# of epochs & 50\\
    & & Batch size & 64\\
    \hline
    \multicolumn{4}{l}{\fontsize{6pt}{8pt}\selectfont\textsuperscript{\textdagger} See \cref{tab:method-hyperparameters}.}\\
    \multicolumn{4}{p{\dimexpr\linewidth-2\tabcolsep\relax}}{\fontsize{6pt}{8pt}\selectfont\textsuperscript{\textdaggerdbl} These are bugs. They should have been different values according to \cref{tab:method-hyperparameters}.}\\
    \end{tabular}
\end{table}

\subsection{Ensembling}

We formed ensembles from the models trained in the hyperparameter tuning.
Models are added to the ensemble in descending order of MRP F1 score on validation dataset (II) until MRP F1 score of the ensemble no longer improves.

For DM and PSD, we simply averaged edge predictions $y_{\text{fw}, i, j}^\text{edge}$ and label predictions $y_{\text{fw}, i, j, c}^\text{label}$, respectively.
On the other hand, the simple average ensembling cannot be applied to UCCA, because number of nodes maybe distinct to each model due to the non-terminal node generation.
Hence, we propose to use a two-step \emph{voting ensemble} for UCCA; for each input sentence, \begin{enumerate*}[label=(\arabic*)]
  \item the most popular pointer sequence is chosen, and
  \item edge and label predictions from the models that outputted the chosen sequence are averaged in the same way as DM and PSD
\end{enumerate*}.

For EDS, we do not explicitly use ensemble learning, but utilize DM graphs from ensembled DM models to reconstruct EDS graphs.
For AMR, we do not use ensembles.

\end{document}